\documentclass{article}
\usepackage{spconf,amsmath,graphicx}
\usepackage{multirow}
\usepackage{multicol}
\usepackage{caption}
\usepackage{subcaption}
\usepackage{enumitem}
\usepackage{balance}
\usepackage[hidelinks]{hyperref}

\usepackage{float}
\title{Image Realness Assessment and Localization with Multimodal Features}
\name{Lovish Kaushik, Agnij Biswas and Somdyuti Paul \thanks{This research was supported by the ANRF grant \tiny{ANRF/ECRG/2024/003084/ENS}}}
\address{Indian Institute of Technology, Kharagpur}
\begin{document}
%
\maketitle
\begin{abstract}
A reliable method of quantifying the perceptual realness of AI-generated images and identifying visually inconsistent regions is crucial for practical use of AI-generated images and for improving photorealism of generative AI via realness feedback during training. This paper introduces a framework that accomplishes both overall objective realness assessment and local inconsistency identification of AI-generated images using textual descriptions of visual inconsistencies generated by vision-language models trained on large datasets that serve as reliable substitutes for human annotations. Our results demonstrate that the proposed multimodal approach improves objective realness prediction performance and produces dense realness maps that effectively distinguish between realistic and unrealistic spatial regions. 
\end{abstract}
\section{Introduction}
\label{sec:intro}
Recent advances in generative AI led to the emergence of models such as ``Stable Diffusion" \cite{stable-diffusion} and ``DALL-E 2" \cite{dalle2}, which enabled creation of highly photorealistic images that are virtually indistinguishable from natural photographic images. The proliferated use of such generative models raises significant concerns related to privacy, security, disinformation, copyright violations and the undermining of human creativity. Consequently, contemporary research has focused on discriminating AI-generated visual media from real ones. However, the complementary task of quantitatively assessing the perceptual realness of AI-generated visual data remains underexplored. Rather than classifying images as real or fake, this problem involves quantifying the extent to which an image is perceived as real to human observers. Such quantification is important for several applications as explained below: \\
\textbullet \ \textbf{Evaluating generative model performance}: due to the stochasticity inherent in most generative models, the outputs of a generative model for the same image or text prompt can differ widely in terms of their verisimilitude. Grading and ranking such outputs thereby requires an objective method of measuring realness. \\
\textbullet \ \textbf{Cross-model selection}: as different generative models yield outputs having varying extents of realness for the same input prompt, several applications can benefit from selecting the best model for a particular prompt using an objective realness metric to rank their outputs. \\
\textbullet \ \textbf{Automated dataset augmentation}: AI-generated visual data can be used to build and augment datasets for domains where there is a paucity of real images due to the difficulty or cost of acquiring them, such as in the case of biomedical and astronomical images, copyrighted visual media, etc. To ensure the reliability of supplementing real data with AI-generated ones, they must be objectively evaluated to reject samples whose perceived realness does not meet acceptable standards. 

Furthermore, localizing unrealistic regions of AI-generated images (AIGI) is also crucial for substituting real images with AIGI , to enhance explainability of generative models and to provide feedback for improving generative models. Yet, this problem remains unaddressed. 

In contrast to prior work, we leverage the human-like perceptual capabilities of state-of-the-art vision–language models (VLMs) for realness quantification and introduce the first dense realness mapping framework for pixel-level interpretability. We present \textbf{RE}alness \textbf{A}ssessment and \textbf{L}ocalization using \textbf{M}ultimodal features (REALM), with the following contributions:
\begin{enumerate}[noitemsep, topsep=0pt]
    \item Augmentation of the existing AIGI datasets with VLM-generated textual descriptors of visual inconsistencies. 
    \item Integration of textual guidance with image features for cross-modal objective realness estimation (CORE). 
    \item A dense realness mapping framework (DREAM) that localizes unrealistic regions for pixel-level interpretability. 
\end{enumerate}

\section{Related Work}
\label{sec:relatedwork}

Several benchmark datasets exist for assessing the perceptual quality of AIGIs \cite{zhang2023perceptual, li2023agiqa, wang2023aigciqa2023}. However, they focus on subjective ratings of text–image alignment and technical distortions that are not specific to generative models, making them unsuitable for realness estimation. The AGIN dataset \cite{agin} consists of 6049 AIGIs from five different generative tasks and  provides annotations of technical, rationality, and overall naturalness scores, where the naturalness score combines both technical and rationality aspects. The RAISE dataset \cite{raise} provides subjective realness ratings for 600 images out of which 480 are AIGIs and 120  are natural images.

Common metrics used to evaluate generative models such as IS \cite{inceptionscore}, FID \cite{fid}, and KID \cite{kid} compare distributions of AIGIs and real images, but cannot grade or rank individual samples. In \cite{agin}, the AGIN dataset was used to develop JOINT, an objective image naturalness model that achieved a Spearman's rank order correlation coefficient (SROCC) of 0.75 for rationality prediction, which is closely related to realness prediction. In \cite{raise}, several baselines for realness assessment were established on the RAISE dataset, with a model trained on ResNet-50 features attaining the best SROCC of 0.6798. 

In \cite{pal4vst} a dataset with subjective annotations of perceptual AIGI artifacts was introduced for training a model for segmenting regions that contained such artifacts. However, the primary objective of this work was to correct small distortions via inpainting, unlike our task of estimating dense realness maps that assign continuous pixel-wise realness scores. Prior works have not addressed the same problem to the best of our knowledge. 

\section{Dataset Construction}
\label{sec:dataset}
\begin{figure}[htb]
\centering
\begin{minipage}{0.15\textwidth}
    \centering
    \includegraphics[width=\linewidth]{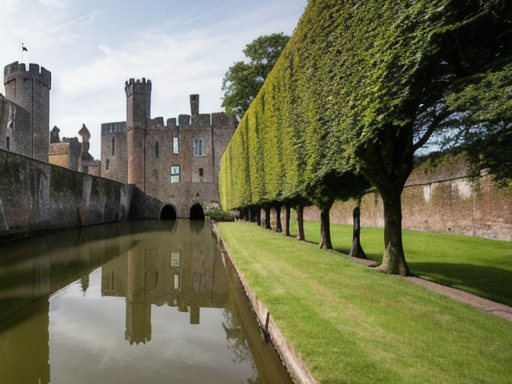}\\
    \vspace{3pt}
    \parbox[c][4cm][t]{\linewidth}{\scriptsize \textbf{LLaVA-1.5:} Yes, the image is somewhat unrealistic as the reflection of the castle in the water is distorted, making it appear as if the castle is floating.\\
    \textbf{GPT-4.1:} Somewhat. The row of trees is unusually squared off at the top with an unnaturally uniform shape, which looks unrealistic for natural tree growth and pruning.}
\end{minipage}
\hfill
\begin{minipage}{0.15\textwidth}
    \centering
    \includegraphics[width=\linewidth]{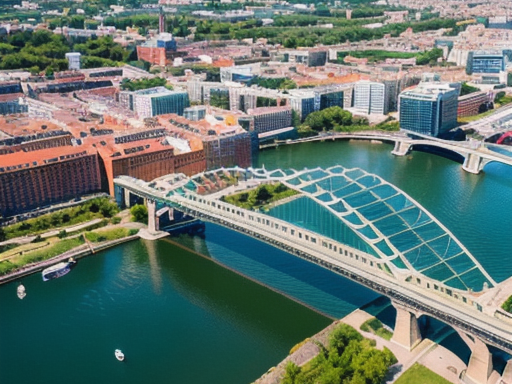}\\
    \vspace{3pt}
    \parbox[c][4cm][t]{\linewidth}{\scriptsize \textbf{LLaVA-1.5:} Yes, the image is somewhat unrealistic as the bridge is shown to be floating above the water.\\
    \\
    \textbf{GPT-4.1:} Yes. The bridge features an unrealistic, irregular structure with warped and inconsistent arches that do not align with real engineering or architectural designs for functional bridges.}
\end{minipage}
\hfill
\begin{minipage}{0.15\textwidth}
    \centering
    \includegraphics[width=\linewidth]{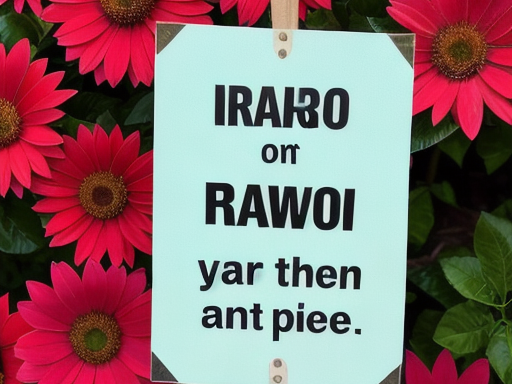}\\
    \vspace{3pt}
    \parbox[c][4cm][t]{\linewidth}{\scriptsize \textbf{LLaVA-1.5:} Yes, the image is somewhat unrealistic as the sign is placed in front of a pink flower, which is not a common sight.\\
    \textbf{GPT-4.1:} Yes. The text on the sign is unrealistic and nonsensical, as it does not form coherent words or sentences, which is unusual for informational or decorative signs.}
\end{minipage}
\caption{Descriptions generated by LLaVA-1.5 and GPT-4.1 on three randomly selected images from the RAISE \cite{raise} dataset.}
\label{fig:dataset}
\end{figure}

As no dataset provides textual descriptions of perceptual realness, we used VLMs to generate such descriptions for images in the RAISE \cite{raise} dataset. Pretrained on large image–text corpora, state-of-the-art VLMs excel at zero/few-shot learning, visual question answering (VQA) and contextualized reasoning tasks. According to \cite{vqa} PaLi-17B \cite{pali}, BLIP-13B 2023 \cite{blip} and LLAVA-1.5 13B \cite{llava} are some of the top performing VQA models. The family of GPT models  \cite{radford2018gpt, radford2019gpt2,brown2020gpt3, openai2023gpt4} also provides compelling multi-modal reasoning capabilities for VQA. A comparison of the textual descriptions generated on a few images by the state-of-the-art open source LLAVA 1.5 \cite{llava} model and GPT-4.1 \cite{openai2023gpt4} in response to the following text prompt is shown in Fig. \ref{fig:dataset}. 
\vspace{0.1cm}
\fbox{\parbox{8.3 cm}{\textbf{Prompt
used : }\textit{Is there anything unrealistic in this image — yes, no, or somewhat? If yes or somewhat, explain in at most 30 words what looks unrealistic, such as a distorted face, uneven object transitions, or any other feature.}}}

In all three images depicted in Fig. \ref{fig:dataset}, GPT-4.1 identifies key inconsistencies such as unusual shapes and nonsensical texts, unlike LLAVA-1.5. Subjective analysis showed GPT-4.1 to be more effective at describing visual inconsistencies in AIGIs. Hence, we used GPT-4.1 to create RAISE-TD, an extension of the RAISE dataset with 600 samples, each comprising an image, its subjective realness MOS, and a realness description obtained with the above prompt.\footnote{The RAISE-TD dataset and the implementation of REALM are available at https://github.com/LOVISH007/REALM}. 

\section{Proposed Method}
\label{sec:methodology}

The proposed REALM framework consists of two components that perform cross-modal objective realness estimation  and dense realness mapping, as elucidated in this section. 

\subsection{Objective Realness Assessment}
\label{subsec:quantification}
\begin{figure}[!htbp]
    \centering
    \includegraphics[width=8cm]{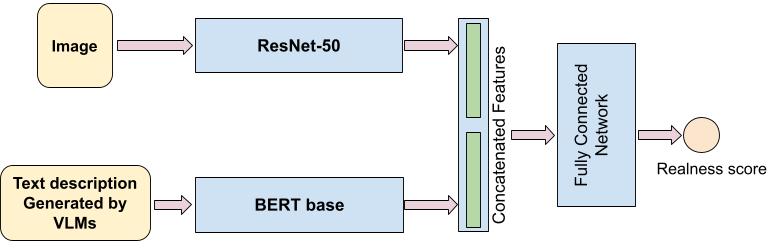 }
    \caption{CORE model for quantifying image realness.}
    \label{fig:model}
\end{figure}

To quantify image realness, we use image–text pairs to predict subjective realness ratings. Image features are extracted with ResNet-50 \cite{he2016resnet} (2048-D) pretrained on ImageNet \cite{deng2009imagenet} and text features with uncased BERT-base \cite{bert} (768-D), pretrained on Wikipedia and BookCorpus \cite{BooksCorpus}. The image and text features are then concatenated and fed to a fully connected (FC) layer having 529 neurons for predicting realness ratings. Both ResNet-50 and BERT are fine-tuned, while the randomly initialized weights of the FC layer are trained to minimize the mean-squared error loss between predicted and ground-truth MOS scores. The proposed model, which we subsequently refer to as \textbf{C}ross-modal \textbf{O}bjective \textbf{R}ealness \textbf{E}stimator (CORE) is illustrated in Fig. \ref{fig:model}.

\subsection{Dense Realness Mapping}

To differentiate unrealistic local regions of AIGIs from realistic regions, we compute continuous-valued dense realness maps that assign high values to pixels in realistic areas and low values to pixels in regions that exhibit visual inconsistencies. Accomplishing this task in a strictly supervised fashion is challenging, as manually annotating unrealistic regions within an image is an onerous task. On the contrary, describing the visual characteristics of an image that make it appear unrealistic is a simpler task for most humans. Based on this insight, we exploit GPT-4.1's \cite{openai2023gpt4} high-level multimodal reasoning ability to accomplish effective unsupervised localization of visual inconsistencies. The descriptions of visual inconsistencies as described in Section \ref{sec:dataset} serve as textual cues for localizing inconsistent regions of AIGIs. 
\begin{figure}[htb]
    \centering  \includegraphics[width=8cm]{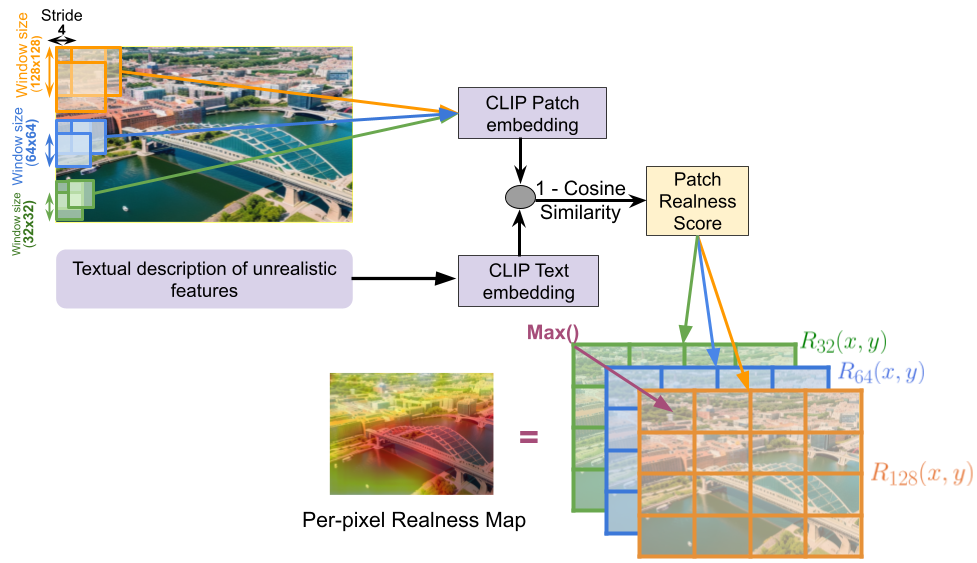}
    \caption{Proposed framework for dense realness mapping}
    \label{fig:localization}
\end{figure}

Fig. \ref{fig:localization} illustrates our localization mechanism, which we henceforth refer to as \textbf{D}ense \textbf{REA}lness \textbf{M}apping (DREAM) Overlapping image patches are extracted, and both patch and text embeddings are obtained using CLIP \cite{clip}. For the $k^{\text{th}}$ patch $P^W_k$ of size $W \times W$, the realness score is computed as:
\begin{equation}
    r^W_{k} = 1 -\frac{\mathbf u^W_{k} \cdot \mathbf{v}}{\|\mathbf u^W_{k}\|\|\mathbf v\|}
\label{eqn:localization}
\end{equation}
where $\mathbf u^W_{k}$ and $\mathbf v$ are 512-dimensional embeddings of patch $P^W_k$ and the text description, respectively. Patches containing unrealistic features yield higher cosine similarity with the text and thus lower realness. 

A sliding-window is used to extract overlapping patches; thus a pixel belongs to multiple overlapping patches. Let $\mathcal P^W_{xy} = \{P^W_k: (x,y) \in P^W_k\}$ be the set of all $W\times W$ patches that contain the pixel at $(x,y)$. The realness scores obtained for the pixel at $(x,y)$ is averaged over all patches in $\mathcal P^W_{xy}$ as:
\begin{equation}
R_W(x,y)=\frac{1}{|\mathcal{P}^W_{xy}|}\underset{\mathcal{P}^W_k}{\sum} r^W_k
\end{equation}

We use multiple patch sizes for multi-scale detection, and the overall realness map selects the best match across multiple scales as $R(x,y) = \max_W R_W(x,y)$. The sliding-window stride controls the granularity of the realness map, where smaller strides produce a more detailed realness map. The values of $R(x,y)$ are normalized between 0-1 to obtain the final realness map.

\section{Experimental Results}
\label{sec:results}

\subsection{Realness Rating Prediction}

The proposed CORE model was trained on an NVIDIA H100 GPU to predict realness MOS ratings using AdamW \cite{AdamW} (learning rate 1e-4) with data augmentation to prevent overfitting.

\subsubsection{Performance on RAISE}

\begin{figure}[htb] 
    \centering
    \includegraphics[width=7cm]{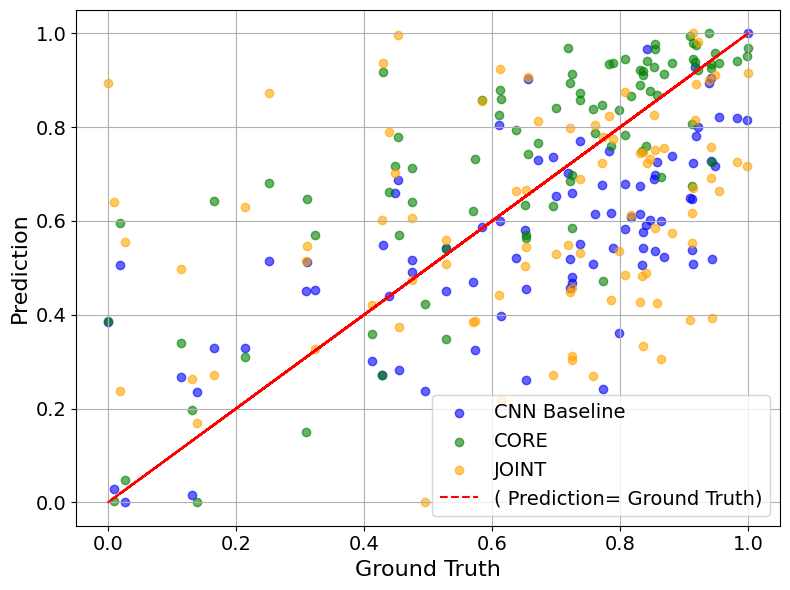} 
    \caption{Scatter plot showing the predictions obtained on the RAISE dataset by the CNN baseline \cite{raise}, JOINT (trained on AGIN) \cite{agin} and CORE. }
    \label{fig:scatter_plot}
\end{figure}

The performance of CORE in terms of SROCC and Pearson's linear correlation coefficient (PLCC) on the 90 images from the test set of RAISE-TD dataset is reported in Table \ref{tab:correlation}. Table \ref{tab:correlation} also reports the performances of the best performing baseline CNN model from \cite{raise} that uses a ResNet-50 backbone pretrained on the AGIN dataset and fine-tuned on the RAISE dataset. The CORE model that uses text and image features for realness estimation achieved an SROCC of 0.7778, significantly outperforming the baseline model proposed in \cite{raise} that estimates realness ratings solely based on image features. The predictions of CORE and CNN baseline against the ground truth MOS ratings is shown in Fig. \ref{fig:scatter_plot}.


\begin{table}[htb]
\centering
\caption{Summary of Realness Prediction Performance}
\label{tab:correlation}
\scalebox{1.0}{
\begin{tabular}{|l|l|c|c|}
\hline
\textbf{Dataset} & \textbf{Model} & \textbf{SROCC} & \textbf{PLCC} \\
\hline
\multirow{3}{*}{RAISE} 
  & CNN baseline \cite{raise}     & 0.6798             & 0.7421             \\
  & Our approach  (CORE)         & \textbf{0.7778}    & \textbf{0.7976}    \\
\hline
\multirow{3}{*}{AGIN} 
  & JOINT \cite{agin}      & \textbf{0.7564} & \textbf{0.7711} \\
  & CNN baseline \cite{raise}     & 0.6314 & 0.6478 \\
  & Our approach (CORE)         & 0.7544 & 0.7639 \\
\hline
\end{tabular}}
\end{table}

\begin{figure*}[htb]
    \centering  \includegraphics[width=18cm]{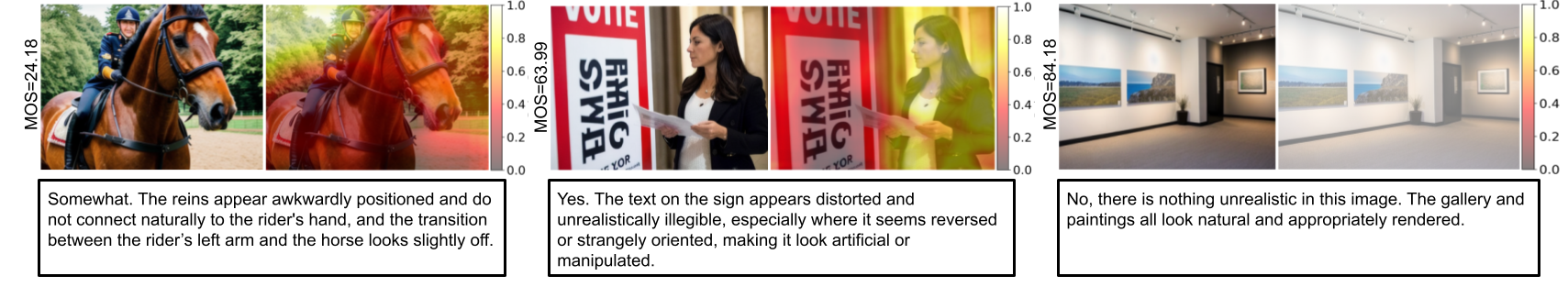}
    \caption{Realness maps of three images computed by DREAM}
    \label{fig:localization_results}
\end{figure*}

\subsubsection{Performance on AGIN}


The JOINT rationality branch trained on AGIN performed poorly on RAISE (SROCC 0.2998, PLCC 0.2776), while the CORE trained on RAISE showed only modest generalization to AGIN (SROCC 0.4249, PLCC 0.3434). The low values of the correlation coefficients suggest poor cross dataset generalization between AGIN and RAISE, which can be attributed to the intrinsic differences in their subjective rating protocol. While RAISE collected overall realness ratings on a continuous scale without any specific guidance, AGIN collected discrete rationality ratings focused on five distinct factors (scene existence, content comprehension, and naturalness of color, layout, and context).

To evaluate CORE on AGIN, we generated GPT-4.1 text descriptions for all the images in AGIN using the prompt specified in Section \ref{sec:dataset}. Since AGIN lacks a fixed test partition, the models were trained \cite{raise} with 5-fold splits (4834 training, 605 test images per fold). The last three rows of Table \ref{tab:correlation} report the average performance of the models under comparison. CORE achieved an SROCC of 0.7544, outperforming the CNN baseline, while JOINT performed slightly better (SROCC 0.7564) owing to its backbone being pretrained on aesthetic assessment and fine-tuned with a SROCC-based loss, unlike CORE.

\subsection{Localization Performance}

We used window sizes of 128, 64, and 32 with a stride of 4 to compute dense realness maps with DREAM. As shown in Fig. \ref{fig:localization_results}, unrealistic regions (e.g., the  horse’s reins, gibberish text) receive low scores, while realistic regions (such as those in the real image with a high MOS rating), receive high scores. To the best of our knowledge, there are no datasets with dense realness annotations, making quantitative evaluation of localization infeasible. However, the qualitative results that we present demonstrate DREAM's value as an interpretable tool for analyzing AIGIs.

\subsection{Ablation study}

We conducted an ablation study to formally analyze the impact of multimodal features in objective realness assessment. The ablation study was performed as follows: \\
\textbullet  \ \textbf{\textit{Image Only Training}}: only the image was provided as training input, with the textual description being replaced with an empty string \\
\textbullet  \ \textbf{\textit{Text Only Training}}: only the textual description was provided as training input, with the image input being replaced with an empty tensor (i.e. tensor of zeroes). \\
\textbullet  \ \textbf{\textit{Text and Image Training}}: both the image and text description were provided as training inputs. 

\begin{table}[htb]
\centering
\caption{Results of Ablation Study}

\scalebox{1.0}{
\begin{tabular}{|l|c|c|}
\hline
\textbf{Inputs} & \textbf{SROCC} & \textbf{PLCC} \\ \hline
Image only (empty description)      & 0.6970 & 0.7874 \\ \hline
Text only (empty image)          & 0.6656 & 0.6301 \\ \hline
Image + Text (CORE)        & \textbf{0.7778} & \textbf{0.7976} \\ 
\hline
\end{tabular}}
\label{tab:ablation}
\vspace{0.1cm}
\end{table}

Combining image and text features yielded significantly higher SROCC and PLCC than using either modality alone, as shown in Table \ref{tab:ablation}. This confirms that the cross-modal learning employed in CORE provides complementary contextual information that is conducive to objective realness assessment.

\section{Conclusion}

We propose REALM, a multimodal framework for image realness quantification and localization that leverages both visual and textual features. Our experiments demonstrate that combining these modalities significantly outperforms unimodal baselines for perceptual realness assessment and enables explainability through dense realness maps that localize unrealistic regions. However, text descriptions generated by GPT-4.1 are often inaccurate, especially in the presence of specific visual characteristics, such as distortions in human faces, which impairs performance. Additionally, GPT-4.1 is expensive to deploy at scale as it is not open-sourced. As future work, we seek to develop a more robust and economical solution by fine-tuning open-source VLMs with human-labeled image realness descriptions. Moreover, as the currently used CLIP model lacks relational context for matching of image patches with the text descriptions, we plan to improve the precision of the realness maps by employing advanced image-text matching approaches that take relational context into account.

\balance
\bibliographystyle{IEEEtran}
\footnotesize{
\bibliography{references}}

\end{document}